\pdfoutput=1

\documentclass[11pt]{article}

\usepackage{acl}

\usepackage{times}
\usepackage{latexsym}

\usepackage[T1]{fontenc}

\usepackage[utf8]{inputenc}

\usepackage{microtype}

\usepackage{hyperref}
\usepackage{url}

\usepackage{graphicx}
\usepackage{amsfonts}
\usepackage{amsmath}
\usepackage{amssymb}
\usepackage{pifont}
\usepackage{bm}
\usepackage{nicefrac}
\usepackage{multirow}
\usepackage{subfigure}
\usepackage[font=small]{caption}
\usepackage{soul}
\usepackage{xcolor}
\usepackage{xspace}
\usepackage{booktabs}
\usepackage{comment}
\usepackage{sidecap}

\newcommand\sennmt{{SenNMT}\xspace}
\newcommand\docnmt{{DocNMT}\xspace}
\newcommand\seninfer{{SenInfer}\xspace}
\newcommand\docinfer{{DocInfer}\xspace}
\newcommand\europarl{Europarl-7\xspace}
\newcommand\iwslt{IWSLT-10\xspace}
\newcommand\otm{En$\rightarrow$Xx\xspace}
\newcommand\mto{Xx$\rightarrow$En\xspace}

\title{Multilingual Document-Level Translation Enables Zero-Shot Transfer From Sentences to Documents}

\author{Biao Zhang$^1$\thanks{~~Work done while Biao Zhang was interning at Google Research.}, Ankur Bapna$^2$, Melvin Johnson$^2$, \\ 
{\bf Ali Dabirmoghaddam$^2$, Naveen Arivazhagan$^2$, Orhan Firat$^2$ } \\
$^1$ School of Informatics, University of Edinburgh \\
$^2$ Google Research \\
\resizebox{\textwidth}{!}{\texttt{b.zhang@ed.ac.uk, \{ankurbpn, melvinp, dabir, navari, orhanf\}@google.com}}
}

\begin{document}
\maketitle
\begin{abstract}

Document-level neural machine translation (\docnmt) achieves coherent translations by incorporating cross-sentence context. However, for most language pairs there's a shortage of parallel documents, although parallel sentences are readily available. In this paper, we study whether and how contextual modeling in \docnmt is transferable via multilingual modeling. We focus on the scenario of zero-shot transfer from \textit{teacher} languages with document level data to \textit{student} languages with no documents but sentence level data, and for the first time treat document-level translation as a transfer learning problem. Using simple concatenation-based \docnmt, we explore the effect of 3 factors on the transfer: the number of teacher languages with document level data, the balance between document and sentence level data at training, and the data condition of parallel documents (genuine vs.\ back-translated). Our experiments on \europarl and \iwslt show the feasibility of multilingual transfer for \docnmt, particularly on document-specific metrics. We observe that more teacher languages and adequate data balance both contribute to better transfer quality. Surprisingly, the transfer is less sensitive to the data condition, where multilingual \docnmt delivers decent performance with either back-translated or genuine document pairs.


\end{abstract}

\section{Introduction}

Recent years have witnessed a trend moving from sentence-level neural machine translation (\sennmt) to its document-level counterpart (\docnmt). \sennmt inevitably suffers from translation errors related with document phenomena~\cite{10.1145/3441691_survey} and delivers obviously inferior performance when compared against human translations and evaluated at a document level~\cite{laubli-etal-2018-machine}. Most efforts on \docnmt aim at improving contextual modeling via dedicated model architectures and/or decoding algorithms~\cite{bawden-etal-2018-evaluating,voita-etal-2019-good,chen-etal-2020-modeling-discourse} and heavily rely on large-scale parallel document resources. Nevertheless, document resources are unevenly distributed across language pairs, with most pairs having little to no such resources.\footnote{Note that we use \textit{language} and \textit{language pair} interchangeably since one side of our parallel data is always English.}


\begin{figure}[t]
    \centering
    \includegraphics[scale=0.65]{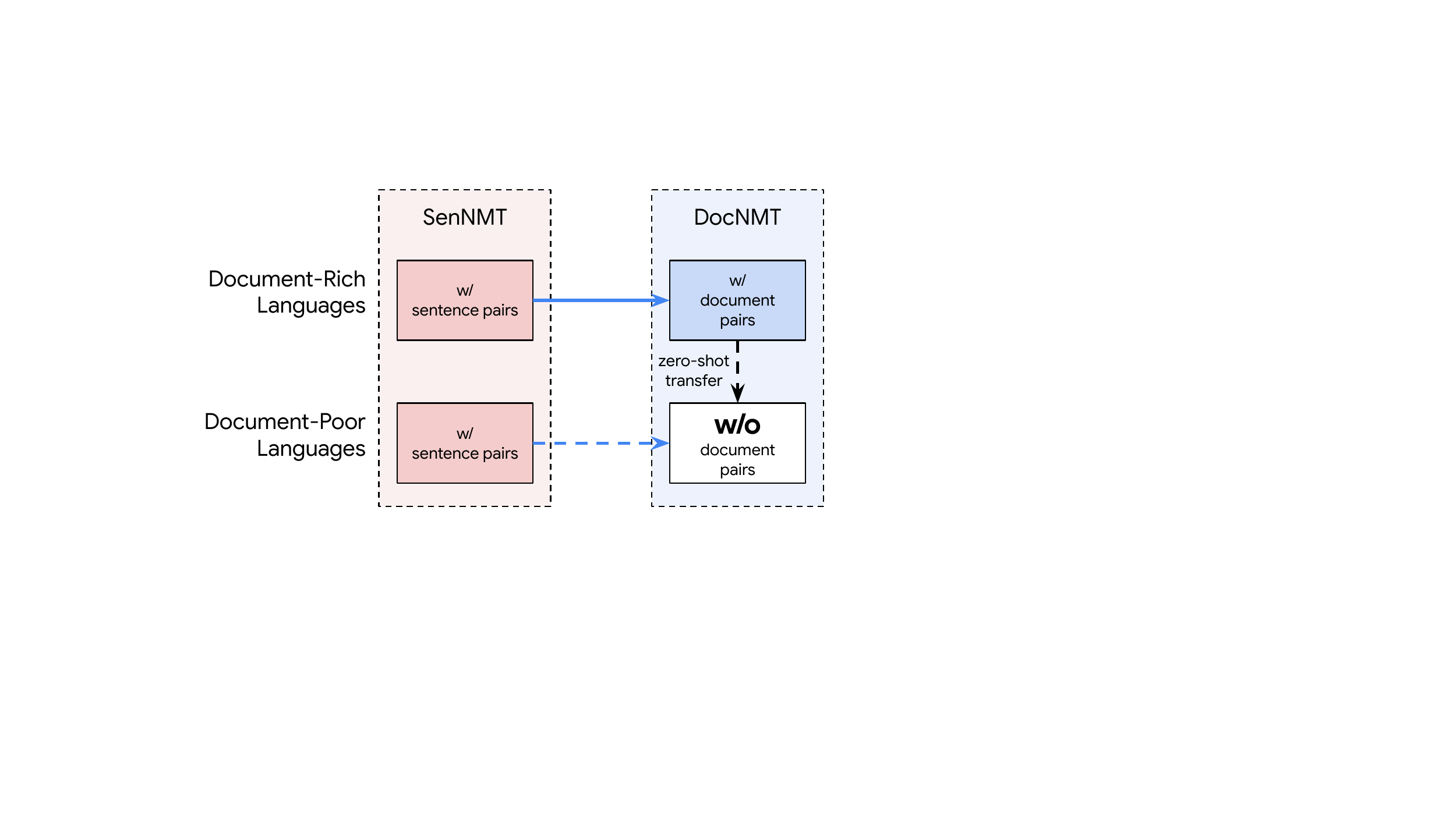}
    \caption{\label{fig:overview} Overview of the focused zero-shot problem for \docnmt. We study transferring contextual modeling from document-rich (teacher) languages to document-poor (student) languages, where only sentence pairs are given for students. The transfer occurs in a multilingual setup, shown by the dashed rectangles. Dashed arrows show the transfer direction.}
    \vspace{-0.5cm}
\end{figure}

One promising way to accommodate languages with varied training data is multilingual modeling, as demonstrated in multilingual \sennmt~\cite{firat-etal-2016-multi,johnson-etal-2017-googles}. 
By sharing parameters across languages, multilingual modeling encourages cross-lingual knowledge transfer, enabling performance improvement and even zero-shot transfer~\cite{aharoni-etal-2019-massively,arivazhagan2019massively,zhang-etal-2020-improving,zhang2021share}. In the context of translation, however, most studies on multilingual transfer center around \sennmt, seldom going beyond sentence-level translation. So far, the question of whether and how document-level contextual modeling can be learned cross-lingually in multilingual \docnmt is still unanswered.

In this paper, we study \textit{zero-shot generalization} for \docnmt{} -- the ability to attain plausible \docnmt quality for some focused (\textit{student}) language pair(s), with only parallel sentences for the student but parallel documents for other (\textit{teacher}) languages in the multilingual mix. The high-level research question we seek to answer is illustrated in Figure \ref{fig:overview}. 

We resort to transfer learning via multilinguality to leverage document resources in teacher languages to help the student languages. We perform our analysis using a simple concatenation based \docnmt, where consecutive sentences are chained into one sequence for translation. We investigate three dimensions extensively to understand the transfer in multilingual \docnmt: 1) the number of languages with document level data (teacher languages), where we simplify our transfer setup to contain either only one teacher language (with N students) or N teachers (with one student); 2) the data balance for parallel documents, i.e. manipulating the ratio of document-level data to sentence-level data during training; and 3) the data condition of parallel documents, where we adopt back-translated parallel documents when only monolingual documents are given in teacher languages or use genuine parallel documents crawled natively. 

We conduct experiments on two publicly available datasets, namely \europarl and \iwslt, covering 6 and 9 languages from/to English respectively. We analyze one-to-many (\otm) and many-to-one (\mto) translation scenarios separately. Following recent work~\cite{ma2021comparison}, we adopt document-specific metrics for evaluation apart from BLEU and support our findings with human evaluations. We also propose a pronoun F1 metric (targeted at gendered pronouns: he/she) for \mto translation, and employ accuracy on contrastive test sets~\cite{bawden-etal-2018-evaluating,muller-etal-2018-large} for \otm translation. Our main findings are summarized below:
\begin{itemize}
    \item Zero-shot transfer from sentences to documents is feasible through multilingual \docnmt modeling, particularly when evaluated with document-specific metrics. This is partially supported by human evaluation.
    \item Transfer quality is strongly affected by the number of teacher languages that use document level data and the data balance for documents. Higher quality is achieved with more teacher languages and adequate document schedule, where the optimal balance varies across scenarios.
    \item Surprisingly, transfer via back-translated documents performs comparable to transfer via genuine parallel documents.
    \item Zero-shot transfer from high-resource document level languages and to low-resource sentence level ones is relatively easier, resulting in better transfer results. 
\end{itemize}

\section{Related Work}

\paragraph{Document-level MT} 
Integrating document-level information meaningfully into NMT is a challenging task, which has inspired research not only on exploring advanced context-aware neural architectures, including simple concatenation-based models~\cite{tiedemann-scherrer-2017-neural,junczys-dowmunt-2019-microsoft,lopes-etal-2020-document}, multi-source models~\cite{jean2017does,bawden-etal-2018-evaluating,zhang-etal-2018-improving}, hierarchical models~\cite{miculicich-etal-2018-document,ijcai2020-551,chen-etal-2020-modeling-discourse}, multi-pass models~\cite{voita-etal-2019-good,yu-etal-2020-better,mansimov2020capturing} and dynamic context models~\cite{kang-etal-2020-dynamic}, to name a few. But it has also motivated the field to revisit the common protocols resorted for evaluation \cite{freitag2021experts}. Despite the hard to measure success, all the above mentioned methods implicitly assume an abundance of document resources and overlook the data scarcity problem. In this study, we adopt the simple concatenation model as our experimental protocol, and leave the exploration of various input formatting options and modelling to future work. Considering the fast changing landscape of the (contextual) MT evaluation, we also provide multiple evaluation metrics including human evaluations, to give a full picture of the phenomena under investigation, while acknowledging the current imperfections of and disagreements on the right way of evaluating MT systems \cite{kocmi2021ship}.


\paragraph{Zero-Shot Transfer via Multilinguality} Multilingual modeling often clusters sentences of similar meaning from different languages within a shared semantic space~\cite{kudugunta-etal-2019-investigating,Siddhant_Johnson_Tsai_Ari_Riesa_Bapna_Firat_Raman_2020}. Such representation space is hypothesized to enable zero-shot transfer, delivering improved performance in many cross-lingual tasks~\cite{eriguchi2018zero,pmlr-v119-hu20b,chi2021infoxlm,ruder2021xtreme}, especially based on large-scale pretrained multilingual Transformers~\cite{devlin-etal-2019-bert,NEURIPS2019_c04c19c2,xue2021mt5}. When it comes to translation, multilingual \sennmt successfully achieves zero-shot translation, transferring sentence-level generation knowledge to language pairs unseen during training~\cite{firat-etal-2016-multi,johnson-etal-2017-googles,gu-etal-2019-improved,arivazhagan2019missing} even in massively multilingual settings~\cite{aharoni-etal-2019-massively,arivazhagan2019massively,zhang-etal-2020-improving,zhang2021share}. Our study extends multilingual \sennmt to multilingual \docnmt and aims at document-level knowledge transfer from languages that have document level data to languages that only have sentence level data. To the best of our knowledge, our study is the first demonstrating the emergence of document-level zero-shot transfer across languages for multilingual machine translation.

\section{Zero-Shot Transfer in Multilingual \docnmt} 

We first formulate the \textit{zero-shot generalization} framework explored in this paper. Given \textbf{N+1} language pairs, we assume that all of them have parallel sentences for training, but only some of them have parallel documents (teachers). Through multilingual training, we study to what degree contextual modeling in document-supervised \docnmt can be transferred to those document-poor (student) languages as in Figure \ref{fig:overview}. Any form of parallel document for student languages is disallowed at training, ensuring that the transfer is measured zero-shot.

\subsection{Multilingual \docnmt}
We employ the concatenation-based method with a $D2D$ structure for \docnmt, where $D$ consecutive sentences in a document are concatenated into one sequence for translation~\cite{junczys-dowmunt-2019-microsoft,sun2020capturing}. Sentence boundary is indicated by a special symbol ``[SEN]''.
We adopt the language token method~\cite{johnson-etal-2017-googles} for multilingual \docnmt, using source and target language token for \mto and \otm translation respectively. Instead of appending this token to the source sequence, we add its embedding to each source word embedding to strengthen the language signal in a document translation setting. 

For \textbf{training}, we adopt a two-stage method: we first pretrain a multilingual \sennmt on sentence level data for all languages; then, we finetune it to obtain multilingual \docnmt on a mix of document level data from teacher languages and sentence level data from student languages. Our analysis requires training a large number of \docnmt models, and the two-stage method saves substantial amounts of computation by sharing the pretrained \sennmt.
For \textbf{evaluation}, we distinguish sentence-level inference (\seninfer) from its document-level counterpart (\docinfer). \seninfer translates sentences separately (out of context), while \docinfer translates $D$ consecutive and non-overlapping sentences in context with each other.\footnote{At decoding phase, the last chunk in a source document can have $<D$ sentences for \docinfer.}

\subsection{Zero-Shot Setup}

We explore three factors for the zero-shot transfer:
\begin{itemize}
    \item \textbf{The number of teacher languages} The source of the transfer comes from teacher languages. Intuitively, both the number of teacher languages and their relevance to student language(s) affect the transfer result. However, exhaustively exploring all possible teacher-student configurations in a multilingual setting will lead to a large search space that expands exponentially with respect to the total number of languages involved. Instead, we simplify our study by exploring two extreme transfer settings, namely \textbf{N21} and \textbf{12N} transfer. The first setting uses N teachers that incorporate document level data with 1 student having sentence level data only, while the second setting has 1 teacher and N students. Note that in either N21 or 12N transfer, there exist N teacher-student configurations, and we report average results over them.\footnote{Note we also include transfer results to individual languages (German and French) in Appendix \ref{app:individual_transfer}.}
    
    \item \textbf{The data balance for parallel documents} When varying the number of teacher languages, the proportion of document data at training also changes. Such imbalance could deeply affect transfer~\cite{arivazhagan2019massively}. To offset this effect, we include the data balance for analysis by controlling the sampling ratio $p$ of documents from 0.1 to 0.9 with a step size of 0.1. Note $p$ is for documents in \textit{all teacher languages}, and the relative proportion among teachers is always retained.
    
    \item \textbf{The data condition of parallel documents} We also study when teacher languages have no parallel documents but only monolingual ones. Methods utilizing monolingual documents for \docnmt vary greatly. Following recent work~\cite{sugiyama-yoshinaga-2019-data,huo-EtAl:2020:WMT,ul-haq-etal-2020-document}, we adopt back-translation (BT) to construct pseudo parallel documents. Note that, for teacher languages, we replace all sentence level training data with pseudo documents rather than mixing them according to our empirical results in Appendix \ref{app:docmodel_propotion}.

\end{itemize}

\section{Experimental Settings}

\paragraph{Datasets} We conduct experiments on two public datasets: \europarl and \iwslt. \europarl is extracted from European Parliament (v10) and has translations between English and $N$=$6$ different languages, including \textit{Czech, German, Finnish, French, Lithuanian and Polish}~\cite{koehn2005epc}. This dataset offers sentence-aligned parallel documents (0.9K$\sim$3.7K documents, 190K$\sim$1.9M sentences) and also monolingual documents (9.7K$\sim$11K documents, 0.65M$\sim$2.28M sentences) for training. For evaluation, we use the WMT dev and test sets~\cite{barrault-EtAl:2020:WMT1} available for each language pair (from 2013 to 2020).
In contrast, \iwslt is collected from TED talks and covers translations between English and $N$=$9$ different languages, including \textit{Arabic, German, French, Italian, Japanese, Korean, Dutch, Romanian and Chinese}~\cite{Cettolo2017}. Unlike \europarl, the distribution of training data over languages in \iwslt is much smoother (uniform). There are $\sim$1.9K sentence-aligned parallel documents with $\sim$240K sentences for each language pair. We further collected about 1K TED talks for each language pair (crawled from Feb 2018 to Jan 2021) as monolingual documents. We use IWSLT17 dev and test sets for evaluation. 
Detailed statistics are given in Appendix~\ref{app:dataset_statistics}.
We preprocess all texts with the byte pair encoding (BPE) algorithm~\cite{sennrich-etal-2016-neural} implemented in the \textit{sentencepiece} toolkit~\cite{kudo-richardson-2018-sentencepiece}, and set the vocabulary size to 32K and 64K for \iwslt and \europarl, respectively.

\paragraph{Model Details} We use the Transformer-base model~\cite{NIPS2017_7181_attention} for experiments with 6 encoder/decoder layers, 8 attention heads and a model dimension of 512/2048. We set $D=5$ for \docnmt. We use Adam~\cite{kingma2014adam} ($\beta_1=0.9$, $\beta_2=0.98$) for parameter update with a learning rate warmup step of 4K and label smoothing rate of 0.1. We apply dropout to residual connections and attention weights with a rate of 0.5 and 0.2, respectively. Other training and decoding details are given in Appendix \ref{app:training_details}.

\paragraph{Back-Translation} Some of our models are trained using back-translated monolingual documents. Back-translations are obtained using bilingual \sennmt (independently for \europarl and \iwslt). To train these models, we halve the BPE vocabulary size as well as the training steps. All other settings are kept as mentioned above.

\paragraph{Evaluation} Following previous work, we use BLEU~\cite{post-2018-call}\footnote{Signature: BLEU+c.mixed+\#.1+s.exp+tok.13a+v.1.4.14} to measure the general translation quality. Document-level BLEU is calculated by counting n-gram at the document level instead of at the individual sentence level~\cite{sun2020capturing}. 

Measuring improvements to document phenomena in translation automatically remains challenging and oftentimes simple surface-based metrics such as BLEU~\cite{laubli-etal-2018-machine} are not sensitive enough. Therefore, we evaluate our model on test sets that focus on such document phenomena. We use the contrastive test sets for En-De~\cite{muller-etal-2018-large} and En-Fr~\cite{bawden-etal-2018-evaluating} which measure a model's ability to distinguish correct from incorrect anaphoric pronoun translations. We include 4 and 1 additional context sentences for En-De and En-Fr contrastive evaluation, respectively. 

Gender bias in translation models has attracted much attention recently~\cite{kuczmarski2018gender,saunders-byrne-2020-reducing}. We expect that contextual information can help to alleviate it. To this end, we introduce gendered pronoun F1 based on the following precision and recall scores to evaluate English translations:
\begin{equation}
    \begin{split}    
        \text{Precision} & = \frac{\sum_{i, g \in \mathcal{G}} \min(C^g_{\textbf{r}_i}, C^g_{\textbf{h}_i})}{\sum_{i, g \in \mathcal{G}} C^g_{\textbf{h}_i}} \\
        \text{Recall} & = \frac{\sum_{i, g \in \mathcal{G}} \min(C^g_{\textbf{r}_i}, C^g_{\textbf{h}_i})}{\sum_{i, g \in \mathcal{G}} C^g_{\textbf{r}_i}}
    \end{split},
\end{equation}
where $\textbf{r}_i$ and $\textbf{h}_i$ denotes the $i$-th gold reference and hypothesis sentence respectively, comprising the gendered pronouns of interest $\mathcal{G}$\footnote{\textit{he, his, him, himself, she, her, hers, herself}.}. $C^g_\textbf{x}$ denotes the count of pronoun $g$ in sentence $\textbf{x}$.

\begin{figure*}[t!]
    \centering
    
    \subfigure{\includegraphics[scale=0.43]{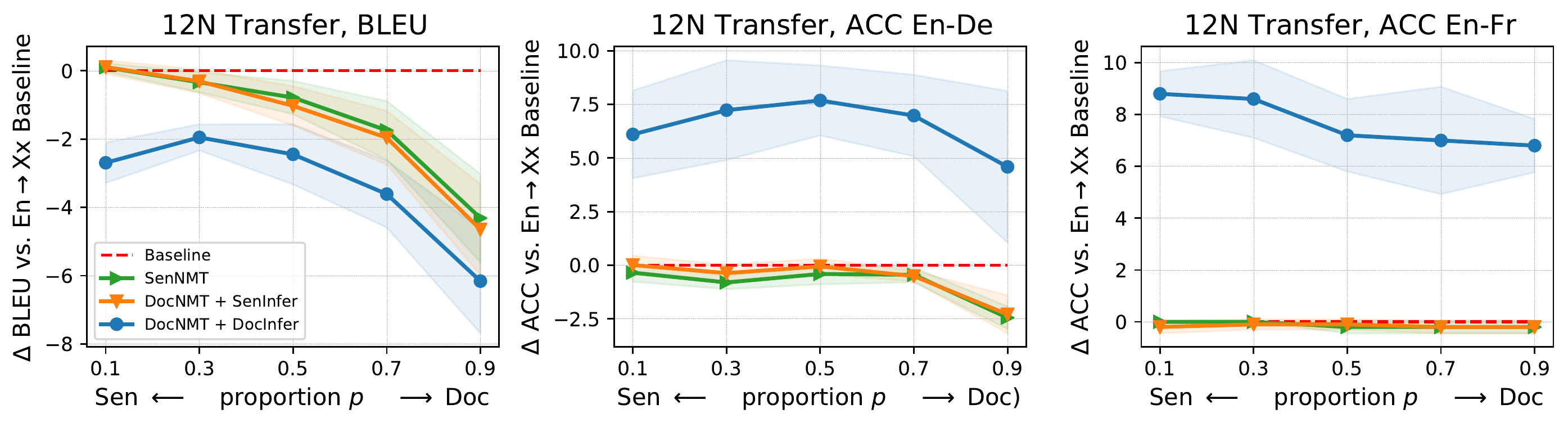}}
    \subfigure{\includegraphics[scale=0.43]{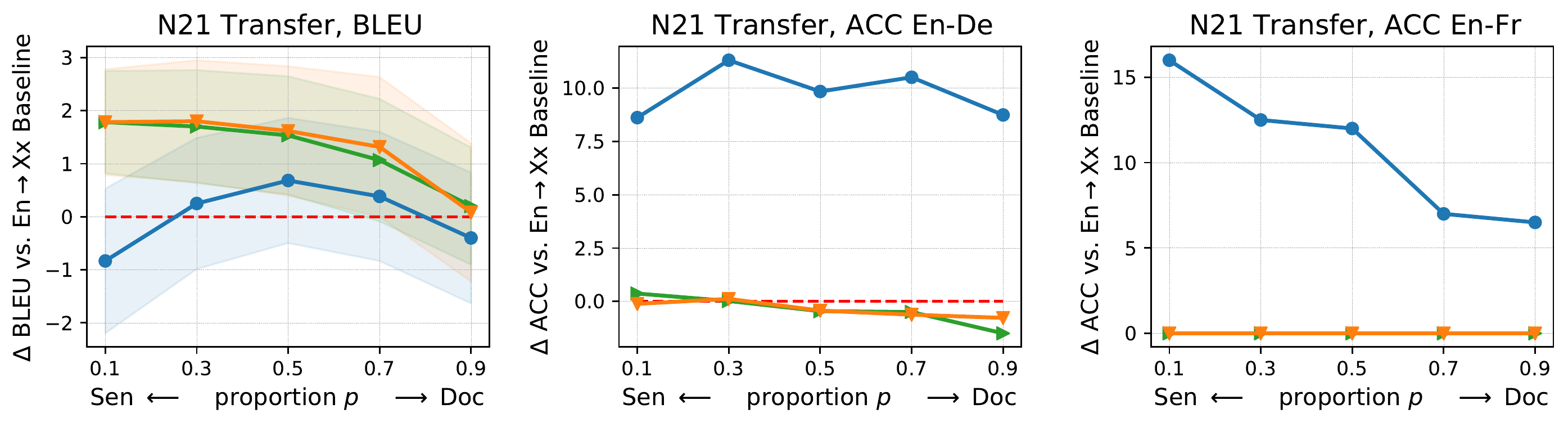}}
    
    \caption{\label{fig:o2m_transfer} Average BLEU and accuracy (ACC) (En-De, En-Fr) in N21 (bottom) and 12N (top) transfer settings for \otm translation on \europarl. Shadow areas denote the standard deviation. $p$ indicates the proportion of documents, and ``$p=0$'': training with student/sentence data alone (\textit{Sen}). ``\textit{Baseline}'': multilingual \sennmt with \seninfer trained with the raw training data. ``\textit{\sennmt}'': the same as Baseline but its sentence-level training data for teacher languages is sampled with a ratio of $p$.
    }
\end{figure*}

\begin{figure*}[t!]
    \centering
    
    \mbox{
    \subfigure{\includegraphics[scale=0.42]{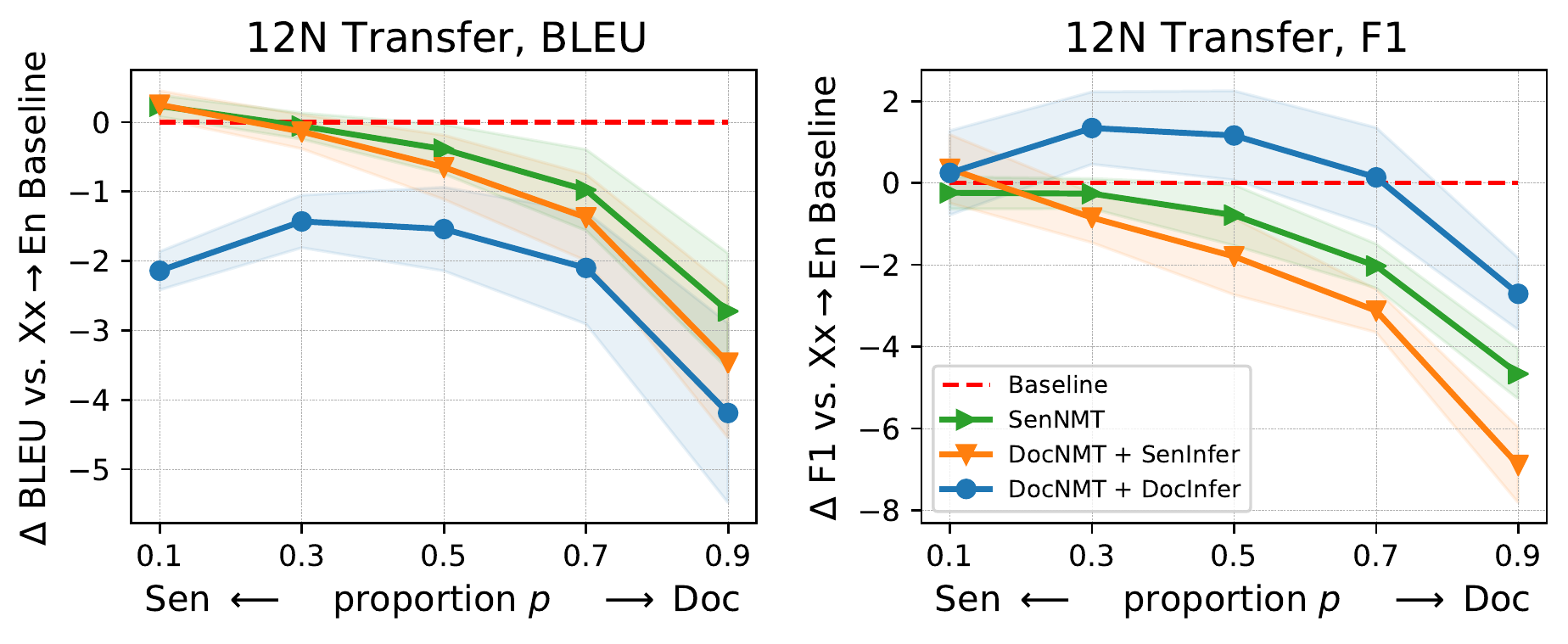}} 
    \subfigure{\includegraphics[scale=0.42]{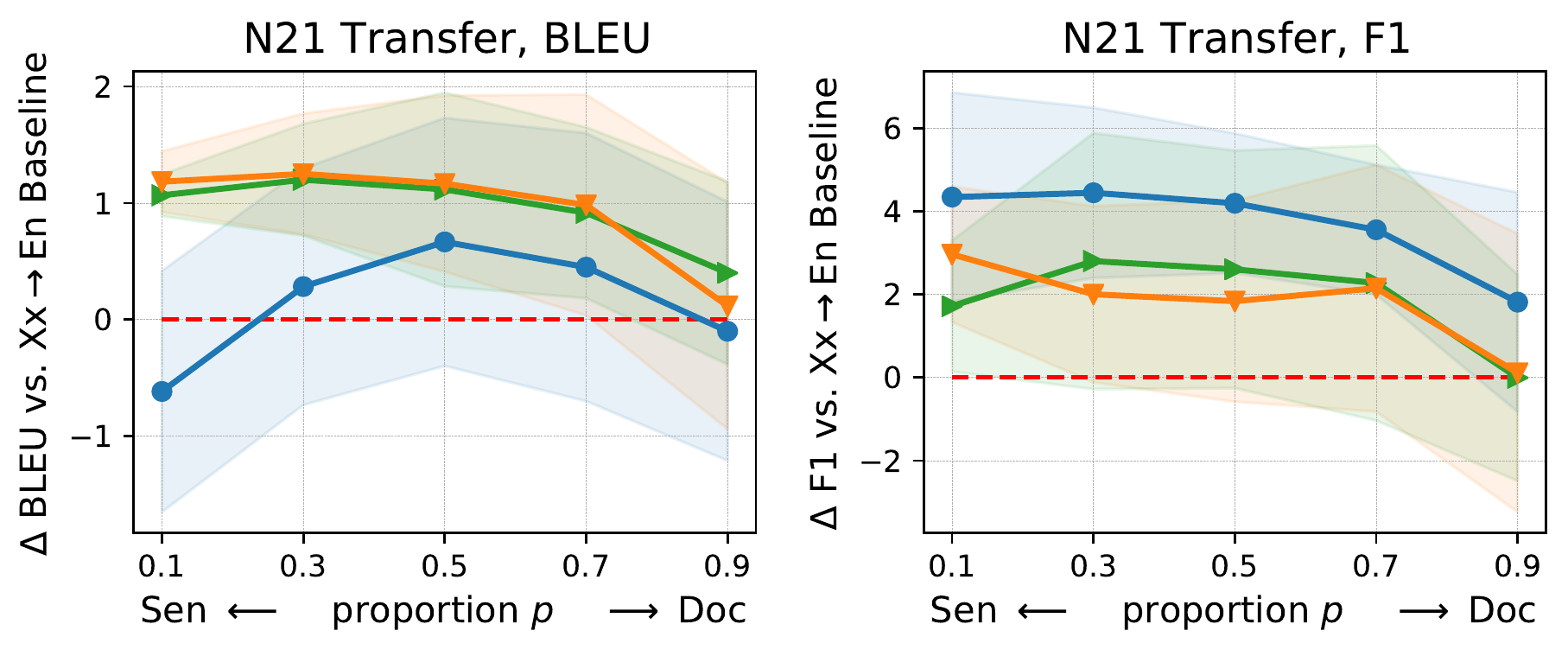}}
    }
    \caption{\label{fig:m2o_transfer} Performance of N21 and 12N transfer as a function of proportion $p$ for \mto translation on \europarl.}
\end{figure*}

Finally, we conduct human evaluation to verify the performance delivered by zero-shot transfer.
We work on En-De, \europarl, where we sample 50 source documents from the test set, and translate them into the target language using the corresponding models and decoding techniques. The translated documents are presented to bilingual human raters who are native in the non-English locale. The raters are asked to evaluate translation qualities while taking the full source document context into account. The raters assign a score in a 0-6 scale to every sentence-translation pair in the document, where 0 and 6 mean nonsense and perfect translations, respectively. For each model, the scores are aggregated across the entire test corpus and the average scores are reported. To ensure a fair diversity of ratings, each rater rates no more than 6 documents per model; an average of 18 raters evaluated each model independently.

\section{Results and Analysis}


\begin{table}[t]
\centering
\small
\begin{tabular}{lcc}
\toprule
Model & \mto & \otm \\
\midrule
\sennmt w/ \seninfer & 22.40 & 18.82 \\
\sennmt w/ \docinfer ($D=2$) & -2.98 & -4.05 \\
\sennmt w/ \docinfer ($D=5$) & -11.7 & -13.0 \\
\bottomrule
\end{tabular}
\caption{\label{tab:sennmt_docinfer} Average BLEU on \europarl for multilingual \sennmt with \seninfer and \docinfer. Rows 2 and 3 represent deltas compared to their Row 1 counterpart. Directly applying \docinfer to \sennmt performs poorly.}
\end{table}

\paragraph{Does \sennmt have the capability of leveraging context?} \textit{Not really!} We put our major analysis on \europarl (N=6, all European languages). Before diving deep into the transfer, we start with analyzing whether \sennmt models trained on sentences alone could generalize to contextual translation. If multilingual \sennmt can be directly used for \docinfer, studying zero-shot transfer would be meaningless. Results in Table \ref{tab:sennmt_docinfer} challenge this possibility: \sennmt results in large quality reduction with \docinfer. We observe that \sennmt produces significantly shorter translations under \docinfer, preferring to translate the first few input sentences. We ascribe such failures to the poor generalization to documents from sentence-level training. 

\begin{figure*}[t]
    \centering
    
    \subfigure{\includegraphics[scale=0.43]{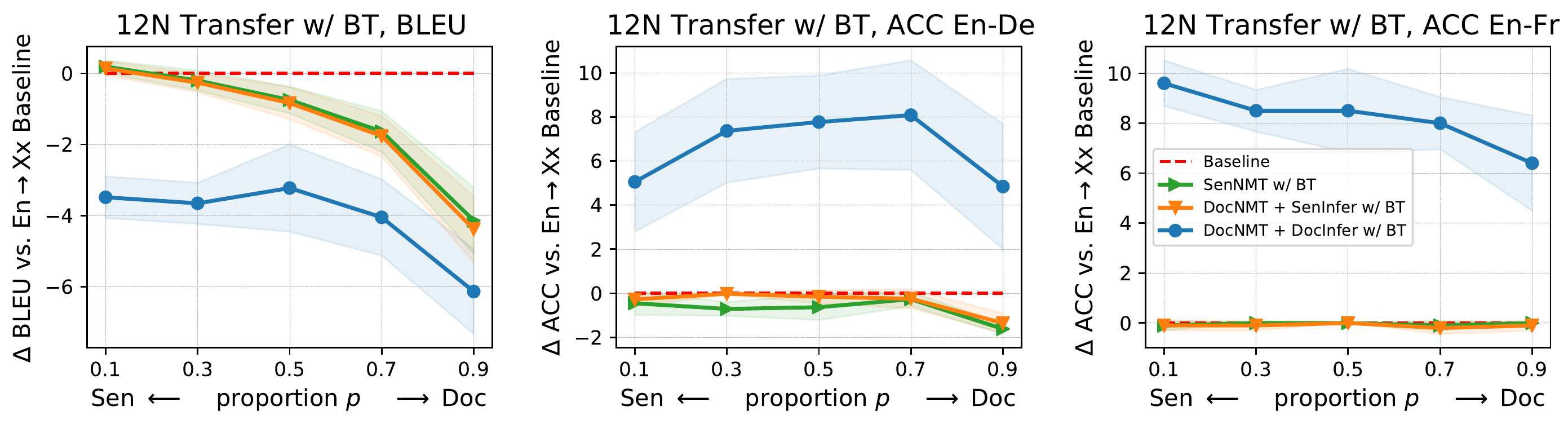}}
    \subfigure{\includegraphics[scale=0.43]{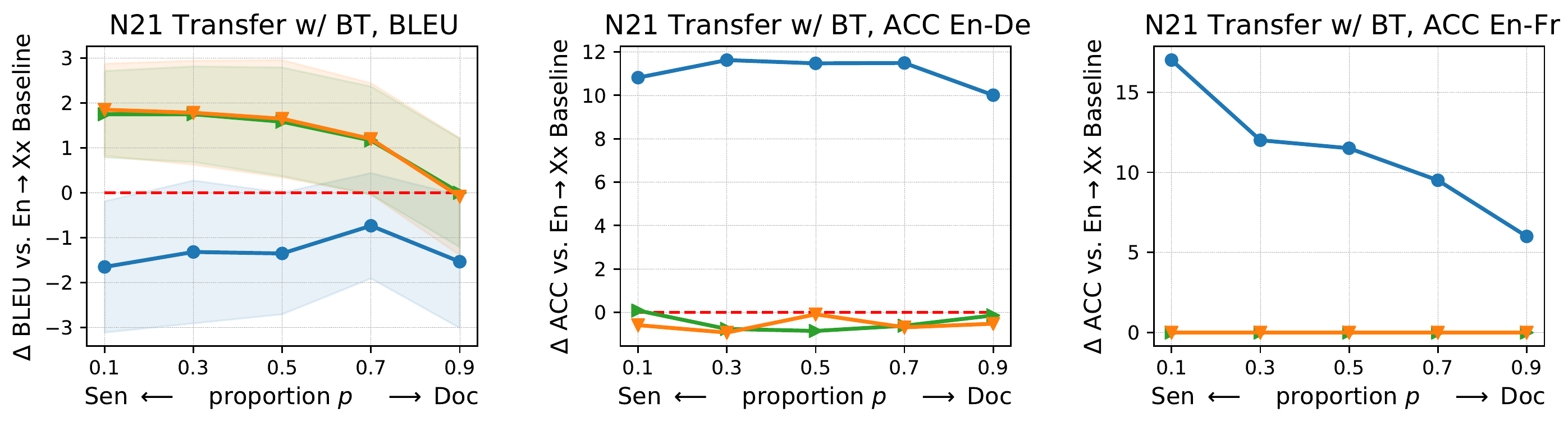}}
    
    \caption{\label{fig:o2m_bt_transfer} Performance of N21 and 12N transfer with back-translated (BT) documents for \otm translation on \europarl.}
\end{figure*}

\begin{figure*}[t]
    \centering
    
    \mbox{
    \subfigure{\includegraphics[scale=0.42]{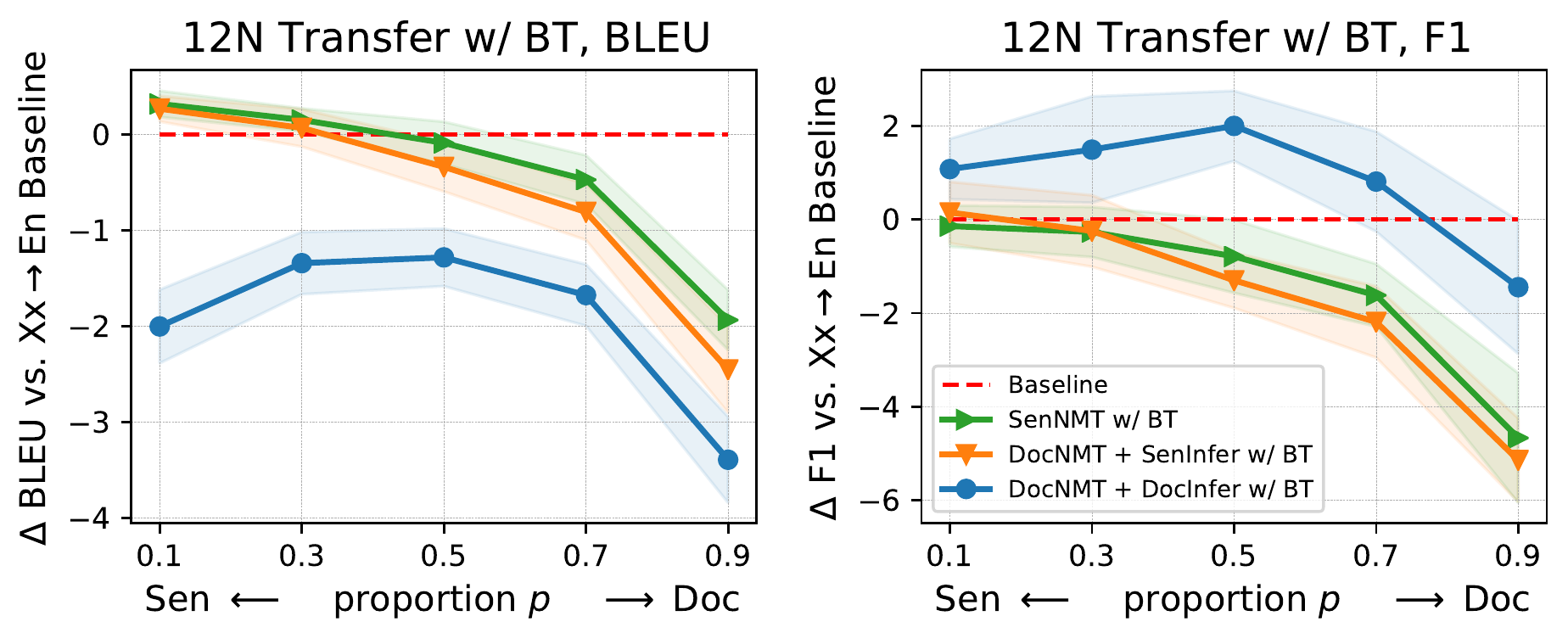}} 
    \subfigure{\includegraphics[scale=0.42]{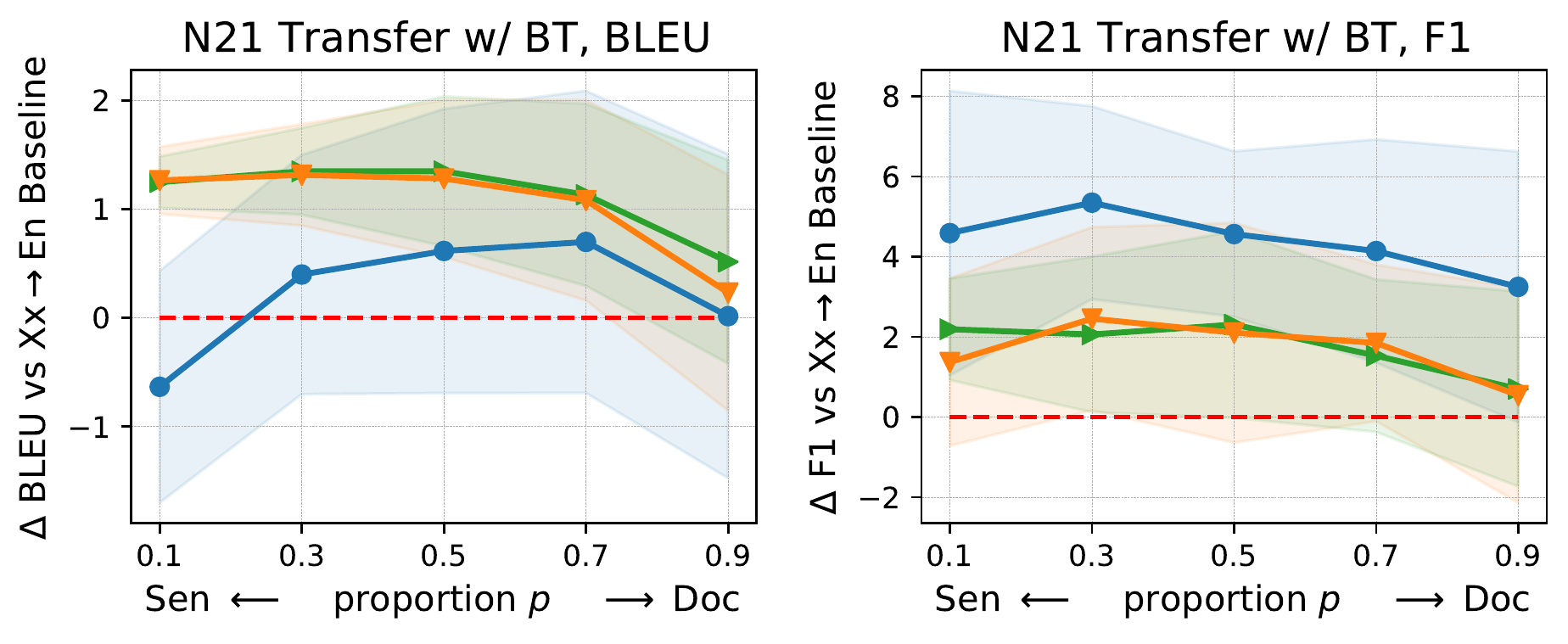}}
    }
    \caption{\label{fig:m2o_bt_transfer} Performance of N21 and 12N transfer with back-translated (BT) documents for \mto translation on \europarl.}
    \vspace{-10pt}
\end{figure*}
\paragraph{Impact of the data balance and the number of teacher languages on zero-shot transfer} 

Figure \ref{fig:o2m_transfer} and \ref{fig:m2o_transfer} summarize the results for \otm and \mto translation, respectively, where we report the average performance paired with the standard deviation over $N$ configurations.\footnote{Note the average results are for transfer directions, not the supervised ones. Each experiment in N21 transfer has only one transfer direction, so we directly report the average over N configurations; by contrast, in 12N transfer, we have N transfer directions, where we first perform average over these N transfer results followed by another average over N configurations. Also note, the average results contains transfer from high/low and similar/distant languages.} Overall, the document-level zero-shot transfer is achievable via multilingual modeling. Transfer-based \docnmt could successfully identify and translate the correct number of input sentences for student languages. With a proper sampling ratio for document-level data, student \docnmt yields better performance than its \sennmt counterpart, especially shown by document-specific evaluations (F1 and ACC).

\begin{table*}[t]
\centering
\small
\begin{tabular}{lcccccccc}
\toprule
\multirow{2}{*}{\mto} & \multicolumn{4}{c}{BLEU} & \multicolumn{4}{c}{F1} \\
\cmidrule(lr){2-5} \cmidrule(lr){6-9}
& High$\rightarrow$ & Low$\rightarrow$ & $\rightarrow$High & $\rightarrow$Low & High$\rightarrow$ & Low$\rightarrow$ & $\rightarrow$High & $\rightarrow$Low \\
\midrule
\docnmt + 12N transfer & \textbf{-1.07} & -1.79 & -1.71 & \textbf{-1.15} & \textbf{+1.73} & +0.95 & +0.73 & \textbf{+1.95} \\
\qquad\qquad w/ BT & \textbf{-1.19} & -1.37 & -1.59 & \textbf{-0.97} & +1.61 & \textbf{+2.38} & +1.36 & \textbf{+2.64} \\
\midrule
\midrule
\multirow{2}{*}{\otm} & \multicolumn{4}{c}{BLEU} & \multicolumn{2}{c}{ACC En-De} & \multicolumn{2}{c}{ACC En-Fr} \\
\cmidrule(lr){2-5} \cmidrule(lr){6-7} \cmidrule(lr){8-9}
& High$\rightarrow$ & Low$\rightarrow$ & $\rightarrow$High & $\rightarrow$Low & High$\rightarrow$ & Low$\rightarrow$ & High$\rightarrow$ & Low$\rightarrow$ \\
\midrule
\docnmt + 12N transfer & \textbf{-1.85} & -2.05 & -2.03 &\textbf{ -1.87} & \textbf{+8.67} & +6.27 & \textbf{+10.25} & +7.50 \\
\qquad\qquad w/ BT & \textbf{-3.29} & -4.03 & -4.39 & \textbf{-2.93} & \textbf{+8.55} & +6.57 & \textbf{+6.61} & +6.19 \\
\bottomrule
\end{tabular}
\caption{\label{tab:high_low_results} Relative performance to multilingual \sennmt baseline when transferring from and into high-resource (High) and low-resource (Low) languages for \otm and \mto translation on \europarl. Results are for \docnmt with \docinfer under 12N transfer; ``High/Low$\rightarrow$'': average results for transferring from high-/low-resource teacher languages to all student languages; ``$\rightarrow$High/Low'': average results for transferring from all teacher languages to high-/low-resource student languages. Transferring from high-resource teacher languages (High$\rightarrow$) and transferring into low-resource student languages ($\rightarrow$Low) show better performance, highlighted in \textbf{bold}.}
\end{table*}

\begin{table*}[t]
\centering
\small
\begin{tabular}{llrrrrr}
\toprule
\multirow{2}{*}{Dataset} & \multirow{2}{*}{Models} & \multicolumn{3}{c}{\otm} & \multicolumn{2}{c}{\mto} \\
\cmidrule(lr){3-5} \cmidrule(lr){6-7}
 & & BLEU & ACC En-De & ACC En-Fr & BLEU & F1 \\
\midrule
\multirow{7}{*}{\europarl}
& \sennmt (Baseline) & 18.82 & 52.14 & 50.00 & 22.40 & 53.81 \\
& \docnmt$^\ddagger$ w/ \seninfer & -0.07 & -0.15 & 0.00 & -0.20 & +0.67 \\
& \docnmt$^\ddagger$ w/ \docinfer & +0.38 & +14.74 & +14.50 & +0.15 & +2.35 \\
\cmidrule{2-7}
& N21 Transfer ($p=0.3$) w/ \docinfer & +0.25 & +11.31 & +12.50 & +0.28 & +4.44 \\
& 12N Transfer ($p=0.3$) w/ \docinfer & -1.95 & +7.23 & +8.60 & -1.43 & +1.34 \\
& N21 Transfer + BT ($p=0.3$) w/ \docinfer & -1.32 & +11.62 & +12.00 & +0.40 & +5.35 \\
& 12N Transfer + BT ($p=0.5$) w/ \docinfer & -3.23 & +7.77 & +8.50 & -1.28 & +2.00 \\
\midrule
\multirow{7}{*}{\iwslt}
& \sennmt (Baseline) & 25.39 & 40.39 & 50.00 & 29.41 & 65.37 \\
& \docnmt$^\ddagger$ w/ \seninfer & -0.77 & +3.19 & 0.00 & +0.26 & +2.42 \\
& \docnmt$^\ddagger$ w/ \docinfer & -0.09 & +15.34 & +8.00 & +0.51 & +4.52 \\
\cmidrule{2-7}
& N21 Transfer ($p=0.3$) w/ \docinfer & +0.01 & +15.74 & +15.00 & +1.10 & +4.40 \\
& 12N Transfer ($p=0.5$) w/ \docinfer & -3.43 & +5.56 & +5.13 & -1.14 & +1.67 \\
& N21 Transfer + BT ($p=0.3$) w/ \docinfer & -1.11 & +13.28 & +18.00 & +1.53 & +3.85 \\
& 12N Transfer + BT ($p=0.5$) w/ \docinfer & -5.32 & +4.19 & +4.88 & -1.71 & +1.58 \\
\bottomrule
\end{tabular}
\caption{\label{tab:final_result_europarl_iwslt} Performance of different models on \europarl and \iwslt. $^\ddagger$: multilingual \docnmt trained on parallel documents from all language pairs. For 12N and N21 transfer, we report one group of results under the approximately optimal proportion $p$. Notice that the results for transfer experiments are averaged over different teacher-student configurations, while those for \docnmt$^\ddagger$ are for one model. We report absolute scores for \sennmt{} but relative scores for the others.}
\vspace{-2pt}
\end{table*}

\textit{Increasing teacher languages improves transfer.} In \otm and \mto translation, we find that N21 transfer performs consistently better than 12N transfer on all metrics. This is reasonable since N21 transfer has $N$ teacher languages, offering richer and more informative sources for transfer. 

\textit{Balancing between document and sentence data matters for transfer.} We also observe that performance changes over the document proportion on all metrics in both 12N and N21 transfer. Applying more or fewer documents during training often hurts zero-shot transfer, indicating a trade-off. Roughly, setting $p$ to 30\%$\sim$50\% delivers good performance (Figure \ref{fig:o2m_transfer} and \ref{fig:m2o_transfer}), although the optimal proportion depends.

\textit{\seninfer underperforms \docinfer on document-specific metrics.}
\docnmt w/ \seninfer performs similarly to \sennmt, and better than \docinfer on BLEU. When evaluating document phenomena, however, \seninfer shows clear insufficiency. This resonates with the findings of~\citet{ma2021comparison}.

\paragraph{Can we achieve zero-shot transfer with monolingual documents?} \textit{Yes.} We next repeat our experiments with BT document pairs. Figure \ref{fig:o2m_bt_transfer} and \ref{fig:m2o_bt_transfer} show that BT performs surprisingly well on document-level zero-shot transfer. We observe almost the same performance pattern compared to training with genuine documents in all settings (\otm and \mto, N21 and 12N transfer and different metrics), although BLEU scores become worse and the optimal proportion also changes. We argue that the target-side genuine context information in BT documents helps contextual modeling~\cite{ma2021comparison}. These results are promising, encouraging further research on exploring monolingual documents for multilingual \docnmt.

\paragraph{Impact of high/low-resource languages on zero-shot transfer.} 
The data distribution of \europarl is highly skewed over languages, with \textit{Cs, Lt, Pl} being relatively low-resource languages while \textit{De, Fi, Fr} being high-resource ones. Studies on multilingual \sennmt have witnessed the transfer from high-resource to low-resource languages~\cite{aharoni-etal-2019-massively,zhang-etal-2020-improving}. We next analyze how this data scale difference affects document-level zero-shot transfer. We mainly explore 12N transfer because of the single transfer source, avoiding interference from other teacher languages.

Table \ref{tab:high_low_results} lists the results. Regardless of the data condition (genuine or BT document pairs), transferring from high-resource teacher languages often outperforms that from low-resource ones. Besides, transferring into low-resource student languages delivers better transfer than into high-resource ones. These suggest that increasing the document data for teacher languages benefits zero-shot transfer.

Note we also provide transfer results from individual languages to De and Fr in Appendix \ref{app:individual_transfer}.

\paragraph{Performance on \europarl and \iwslt} We summarize the main results on both datasets in Table \ref{tab:final_result_europarl_iwslt}. Although \iwslt (N=9) includes more (distant) languages and distributes quite differently over languages, the results on \iwslt resemble those on \europarl. On both datasets, we observe that transfer, both 12N and N21, yields very positive results, particularly with document-specific metrics. Unlike \europarl, BT-based transfer performs much worse than models trained on genuine document pairs on \iwslt. We ascribe this to the data scarcity, where only very small-scale monolingual documents are used for BT in \iwslt. This also reinforces our observation that more document resources benefits zero-shot transfer.


\begin{table}[t]
\centering
\small
\begin{tabular}{lc}
\toprule
\multirow{1}{*}{Models} & \multicolumn{1}{c}{Human Rating ($\uparrow$)} \\
\midrule
Reference & \textbf{4.96} \\
\midrule
\sennmt (Baseline) & 3.31 \\
\docnmt$^\ddagger$ w/ \seninfer & 3.60 \\
\docnmt$^\ddagger$ w/ \docinfer & \textbf{3.84} \\
N21 Transfer w/ \docinfer & 3.46 \\
12N Transfer w/ \docinfer & 2.78 \\
N21 Transfer + BT w/ \docinfer & 3.18  \\
12N Transfer + BT w/ \docinfer & 2.72 \\
\midrule
\end{tabular}
\caption{\label{tab:human_evaluation} Document-level human ratings ($\uparrow$) for En-De on \europarl. We evaluate the best system indicated by document-level metrics (ACC En-De) for 12N and N21 transfer. We randomly sample 50 documents for human evaluation. Ratings are on a 0-6 scale; higher scores mean better quality.}
\end{table}

\section{Discussion}

Apart from automatic evaluation, we also offer human evaluation on En-De. We choose En-De as its WMT20 test set is intentionally constructed for \docnmt evaluation. Table \ref{tab:human_evaluation} lists the results. 


We observe that zero-shot transfer matches and even surpasses \sennmt through N21 transfer, but fails with 12N transfer, although accuracy improvements on contrastive test sets show that both transfers are better than \sennmt. We conjecture that these contrastive test sets only target a limited number of document phenomena and thus can't fully reflect the overall translation quality and represent human preference. These numbers verify the feasibility of document-level zero-shot transfer through multilinguality. Besides, we find that genuine parallel documents benefit the transfer slightly more than BT-based pseudo ones, and that the supervised \docnmt reaches the best result under \docinfer.

We surprisingly find that \docnmt with \seninfer yields very competitive performance, although no contextual information is used for decoding. We also observe that such decoding tends to produce longer translations than \sennmt despite using the same decoding hyperparameters. This behaviour should be shaped by the fact that \docnmt is biased towards long concatenated target references. This partially agrees with the recent argument that context improves \docnmt with some sort of regularization rather than teaching the model to deal with context~\cite{kim-etal-2019-document}. On the other hand, this challenges how to properly evaluate \docnmt.

\begin{table}[t]
\centering
\small
\begin{tabular}{lcc}
\toprule
Models & ACC En-Fr & ACC En-De \\
\midrule
\sennmt w/ \seninfer & 50.00 & 52.00 \\
\sennmt w/ \docinfer & \textbf{58.50}$^\star$ & 50.80 \\
\midrule
\docnmt w/ \seninfer & 50.00 & 51.90 \\
\docnmt w/ \docinfer & \textbf{64.50}$^\dagger$ & \textbf{66.80}$^\dagger$ \\
\bottomrule
\end{tabular}
\caption{\label{tab:problem_with_contrastive} Applying \docinfer and \seninfer to \docnmt and \sennmt for contrastive evaluation. Models are trained on \europarl. $^\star/^\dagger$: significant at $p<0.05/0.01$.}
\vspace{-15pt}
\end{table}

Another observation is that applying \docinfer to \sennmt delivers a significant accuracy improvement on En-Fr contrastive test set (+8.5\%, Table \ref{tab:problem_with_contrastive}), but slightly worse results on En-De. To accurately recognize the correct translation in these test sets, models need to leverage context. Such improvement might suggest that \sennmt has some limited capability of contextual modeling, but might just reflect the instability of small-scale test sets (only 200 cases in En-Fr test set, indicating a radius of around 7\% for the 95\% confidence interval). To some extent, this devalues the improvement achieved by 12N transfer as shown in Table \ref{tab:final_result_europarl_iwslt}, but strengthens the success of N21 transfer (often $>$9\% gains).  

\section{Conclusion and Future Work}

This paper studies the variables playing role in achieving zero-shot document-level translation capability for languages that only have sentence level data (students), through multilingual transfer from languages that have access to document level data (teachers). We make the first step in this direction by extensively exploring properties of transfer by investigating three different variables. Our experiments on \europarl and \iwslt confirm the feasibility, where we discover that increasing document-supervised teacher languages thereby increasing the document training data size, adequately balancing between document and sentence data at training, and leveraging monolingual documents via back-translation all benefit zero-shot transfer in varying degrees. The transferability of contextual modeling in \docnmt demonstrates the potential of delivering multilingual \docnmt with limited document resources.

Along with the success of document-level zero-shot transfer, problems with accurately estimating the document-level translation become challenging. BLEU often fails to capture document phenomena, while contrastive test sets only cover few document-level aspects. Neither perfectly correlates with human evaluation. Besides, whether the gains really come from contextual modeling is still unclear. Our human evaluation shows some preference to \docnmt with \seninfer where context is not used for decoding at all. Designing better evaluation protocols (either automatic or human) is again confirmed to be critical. Besides, performing analysis beyond 12N and N21 transfer deserves more effort and it is an interesting and plausible future direction to analyze how language similarity affects the transfer.

\section*{Acknowledgements}

We thank the reviewers for their insightful comments. 
We want to thank Macduff Hughes and Wolfgang Macherey for their valuable feedback. 
We would also like to thank the Google Translate team for their constructive discussions and comments. 


\bibliography{emnlp2021}


\appendix

\begin{table*}[ht!]
\centering
\small
\begin{tabular}{lccccccccc}
\toprule
\multirow{2}{*}{\begin{tabular}{@{}c@{}}Language \\ (Pair) \end{tabular}}
  & \multicolumn{2}{c}{Train-Para} & \multicolumn{2}{c}{Train-Mono} & \multicolumn{2}{c}{Dev} & \multicolumn{2}{c}{Test} \\
  \cmidrule(lr){2-3} \cmidrule(lr){4-5} \cmidrule(lr){6-7} \cmidrule(lr){8-9}
  & \#Sent & \#Doc & \#Sent & \#Doc & Source & \#Doc & Source & \#Doc \\
\midrule
\europarl \\
\midrule
En-Cs & 192K & 901 & 658K & 9759 & WMT19 & 123 & WMT20 & 130 (102) \\ 
En-De & 1.81M & 3394 & 2.10M & 10155 & WMT19 & 123 (145) & WMT20 & 130 (118) \\
En-Fi & 1.82M & 3569 & 2.00M & 10161 & WMT18 & 132 & WMT19 & 123 (134) \\
En-Fr & 1.90M & 3676 & 2.07M & 10304 & WMT13 & 52 & WMT14 & 176 \\
En-Lt & 189K & 901 & 668K & 9740 & WMT19 & 130 & WMT19 & 62 (76) \\
En-Pl & 191K & 901 & 694K & 9775 & WMT20 & 128 & WMT20 & 63 (62) \\
En & - & - & 2.28M & 11109 & - & - & - & - \\
\midrule
\iwslt \\
\midrule
En-Ar & 232K & 1907 & 107K & 1316 & \multirow{9}{*}{\begin{tabular}{@{}c@{}}IWSLT17 \\ Dev10 \end{tabular}} & 19 & \multirow{9}{*}{\begin{tabular}{@{}c@{}}IWSLT17 \\ Tst17 \end{tabular}}  & 12 \\
En-De & 206K & 1705 & 41K & 466 & & 19 & & 10 \\
En-Fr & 233K & 1914 & 119K & 1300 & & 19 & & 12 \\
En-It & 249K & 1902 & 89K & 942 & & 19 & & 10 \\
En-Ja & 223K & 1863 & 26K & 1037 & & 19 & & 12 \\
En-Ko & 230K & 1920 & 132K & 1153 & & 19 & & 12 \\
En-Nl & 253K & 1805 & 52K & 501 & & 19 & & 10 \\
En-Ro & 237K & 1812 & 64K & 761 & & 19 & & 10 \\
En-Zh & 231K & 1906 & 108K & 1283 & & 19 & & 12 \\
En & - & - & 136K & 1445  & - & - & - & - \\
\bottomrule
\end{tabular}
\caption{\label{tab:dataset} Statistics of train, dev and test data for \europarl and \iwslt. Numbers in the bracket are for the reversed evaluation direction. ``\textit{Para}'': parallel corpus; ``\textit{Mono}'': monolingual corpus; ``\textit{\#Sent/\#Doc}'': number of sentences/documents.}
\end{table*}

\section{Data Statistics}\label{app:dataset_statistics}

Table \ref{tab:dataset} shows the statistics for \europarl and \iwslt. Compared to \iwslt, \europarl includes fewer languages, but with higher quantity and more uneven distribution.

\section{Model Training and Decoding Settings}\label{app:training_details}

We pretrain multilingual \sennmt for 100K and 300K steps on \iwslt and \europarl respectively, and adopt extra 20K finetuning steps for multilingual \docnmt. We train all models (\sennmt \& \docnmt) with a fixed batch size of 1280 samples, and schedule the training data distribution over language pairs according to the sentence-level statistics (without oversampling, and this also applies to \docnmt). All such measures aim to ensure a fair comparison between \sennmt and \docnmt. For training, we truncate sequences with length limit of 100 and 512 for \sennmt and \docnmt separately. We average last 5 checkpoints for evaluation. Beam search is used for decoding with a beam size of 4 and length penalty of 0.6. During decoding, we disable the generation of the end-of-sentence symbol for \docinfer until the model outputs the correct number of target translations.

\section{Impact of Back-Translated Documents on Translation}\label{app:docmodel_propotion}

\begin{figure*}[ht]
    \centering
    \subfigure{\includegraphics[scale=0.43]{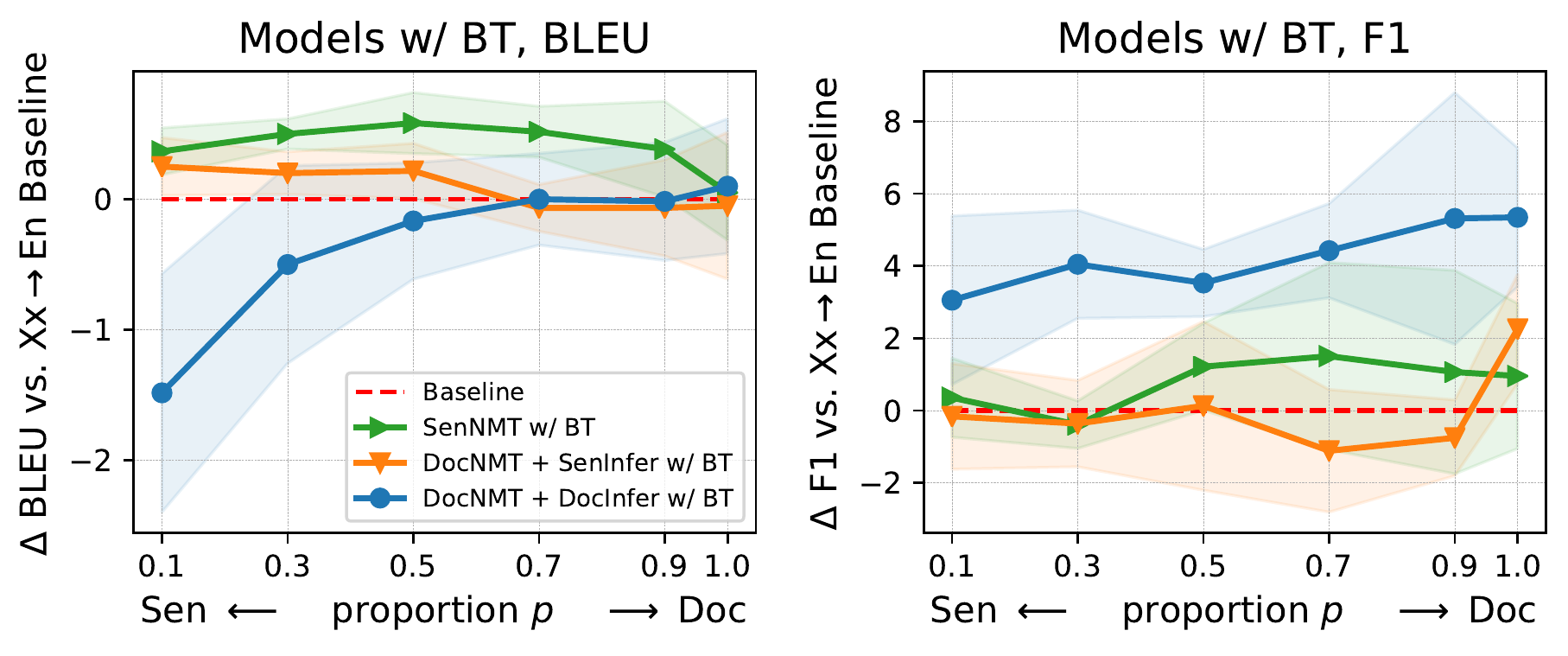}} 
    \subfigure{\includegraphics[scale=0.43]{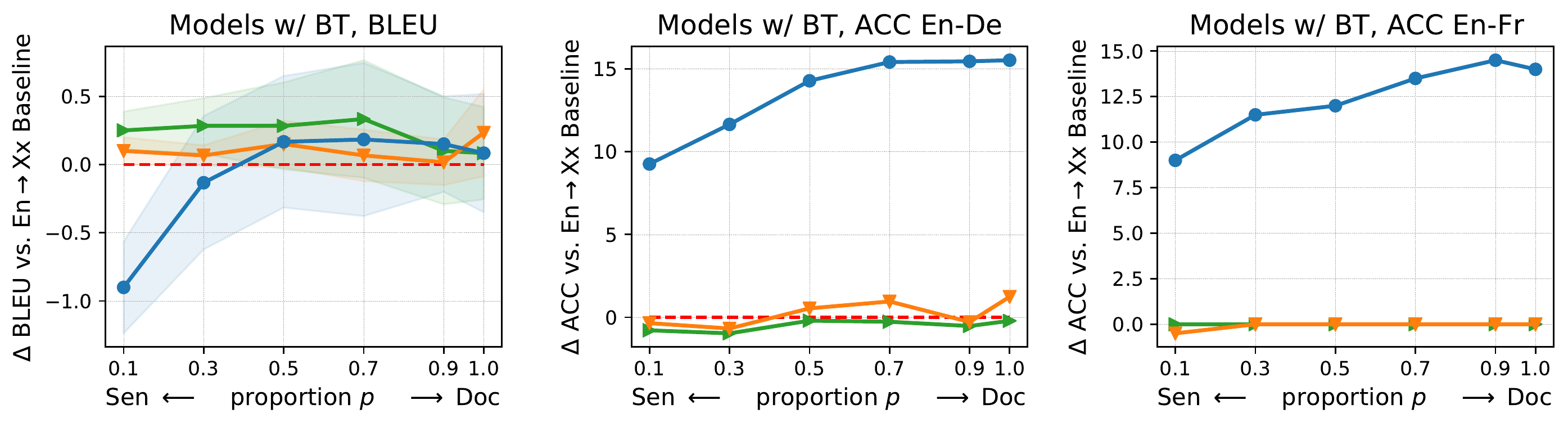}}
    \caption{\label{fig:bt_impact} Translation performance as a function of proportion $p$ with \textbf{back-translated documents} for \otm (bottom) and \mto (top) translation on \europarl. This is fully supervised multilingual \docnmt, where pseudo documents are used for all languages. \textit{Also, note $p$ denotes the proportion of documents for each language, rather than teacher languages.}}
\end{figure*}

\begin{figure*}[ht]
    \centering
    \subfigure{\includegraphics[scale=0.43]{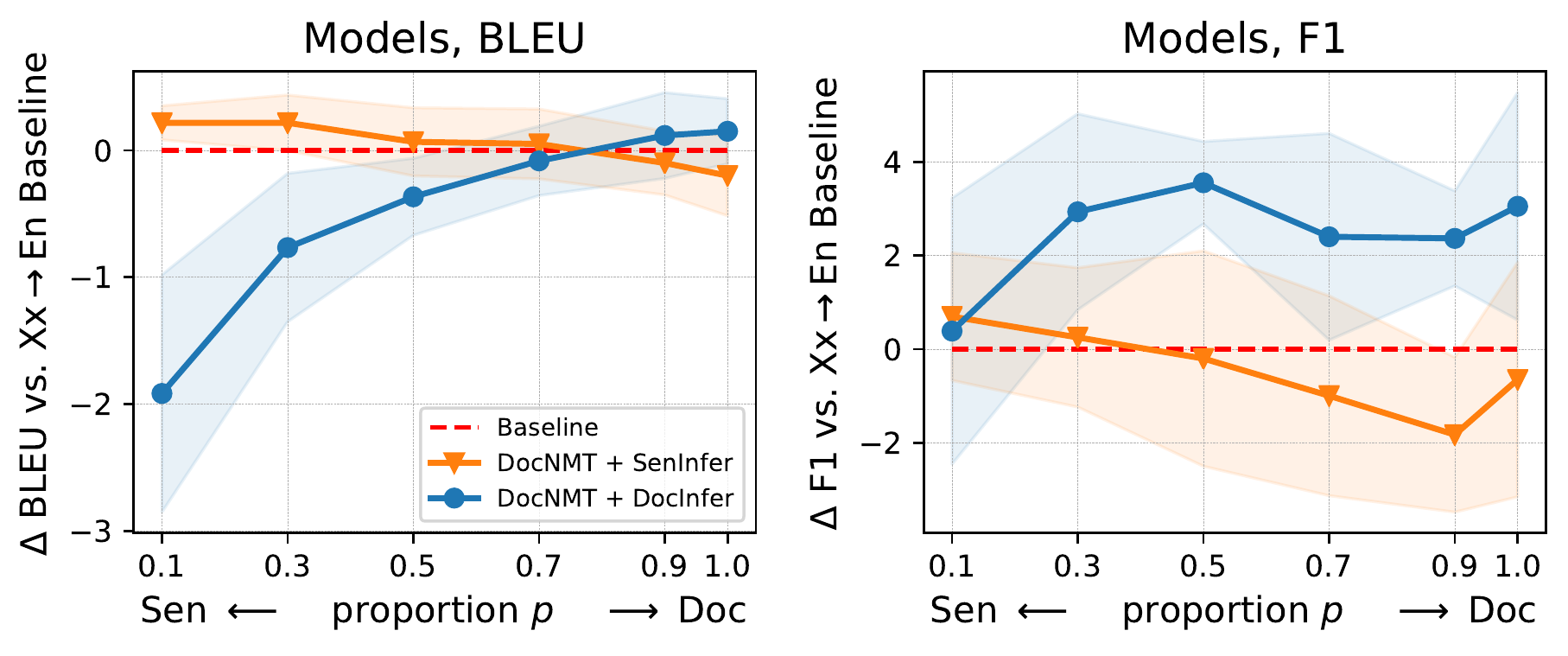}} 
    \subfigure{\includegraphics[scale=0.43]{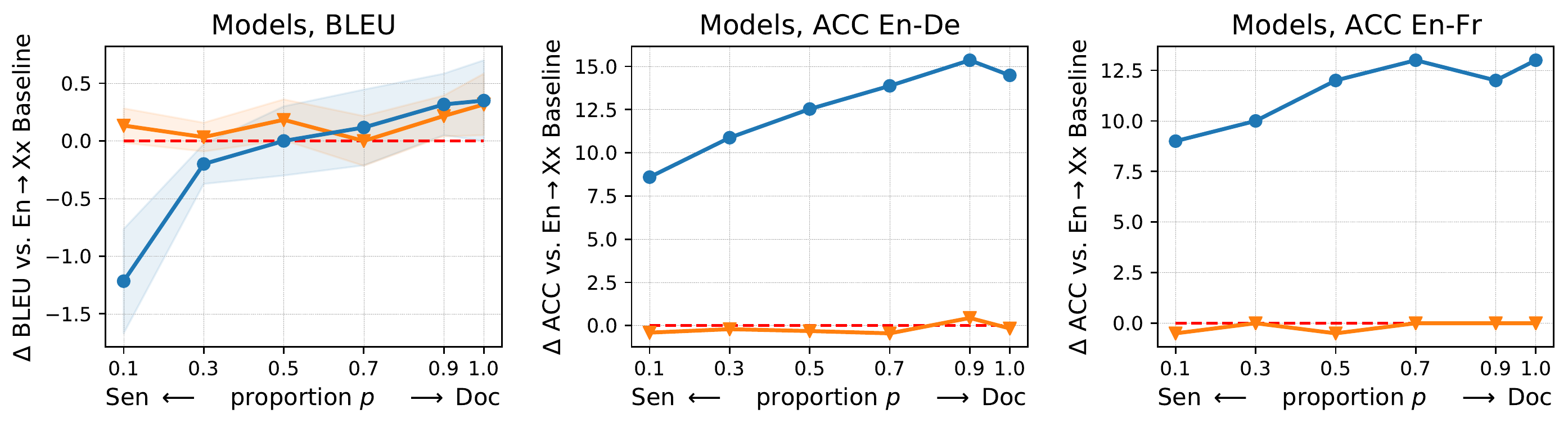}}
    \caption{\label{fig:doc_impact} Translation performance as a function of proportion $p$ with \textbf{genuine parallel documents} for \otm (bottom) and \mto (top) translation on \europarl. Other settings follow Figure \ref{fig:bt_impact}.}
\end{figure*}

The back-translated documents belong to extra training data. How to mix them with the genuine sentence pairs during training is questionable. Before further study, we first explore the impact of these documents on translation. 

Specifically, we sample $p\%$ BT documents for \textbf{each language} during training with the rest ($1-p$\%) being the original sentence pairs to testify the sensitivity of translation performance to $p$. Note the proportion $p$ here differs from the one used in our main paper (where $p$ denotes the proportion of parallel documents in \textbf{all teacher languages} to parallel sentences in student languages).

Figure \ref{fig:bt_impact} shows that larger $p$ generally yields better performance over all settings, similar to the results on genuine parallel documents as in Figure \ref{fig:doc_impact}. Therefore, we replace all sentence pairs in teacher languages with the corresponding BT documents in our analysis.


\section{Transfer Results From Individual Languages to De/Fr}\label{app:individual_transfer}

We mainly report average results over all transfer directions in the paper. Below we also show the transfer from individual languages to De and Fr on \europarl. Note the performance at language level is much noisy. We observe that different teacher languages yield slightly different transfer behaviors and transferring to Fr looks more promising.

\begin{figure*}[ht]
    \centering
    \subfigure{\includegraphics[scale=0.43]{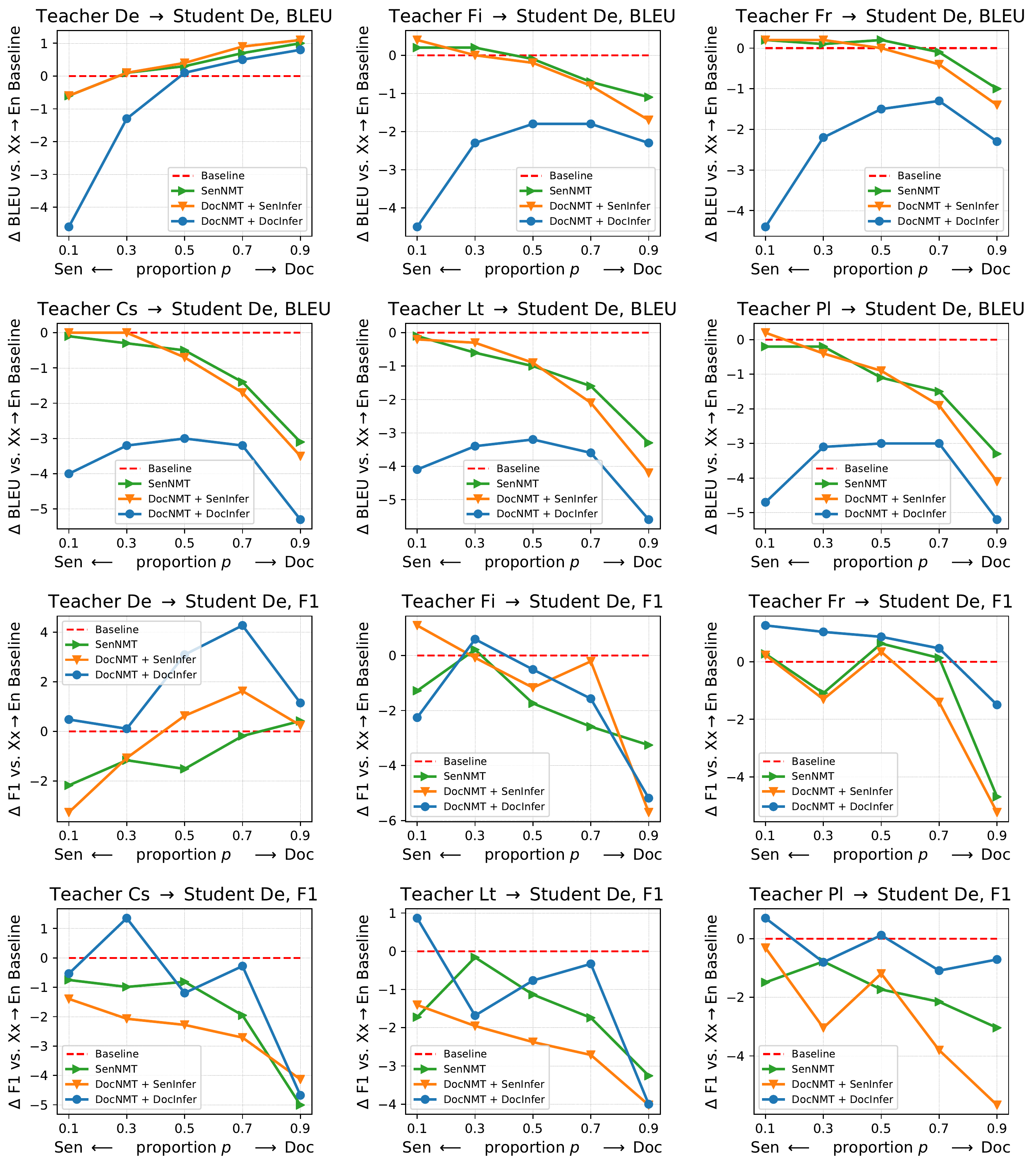}} 
    \caption{\label{fig:individual_transfer_to_de_m2o} Transfer performance from individual languages to De as a function of proportion $p$ with genuine parallel documents for \textbf{\mto} translation on \europarl.}
\end{figure*}

\begin{figure*}[ht]
    \centering
    \subfigure{\includegraphics[scale=0.43]{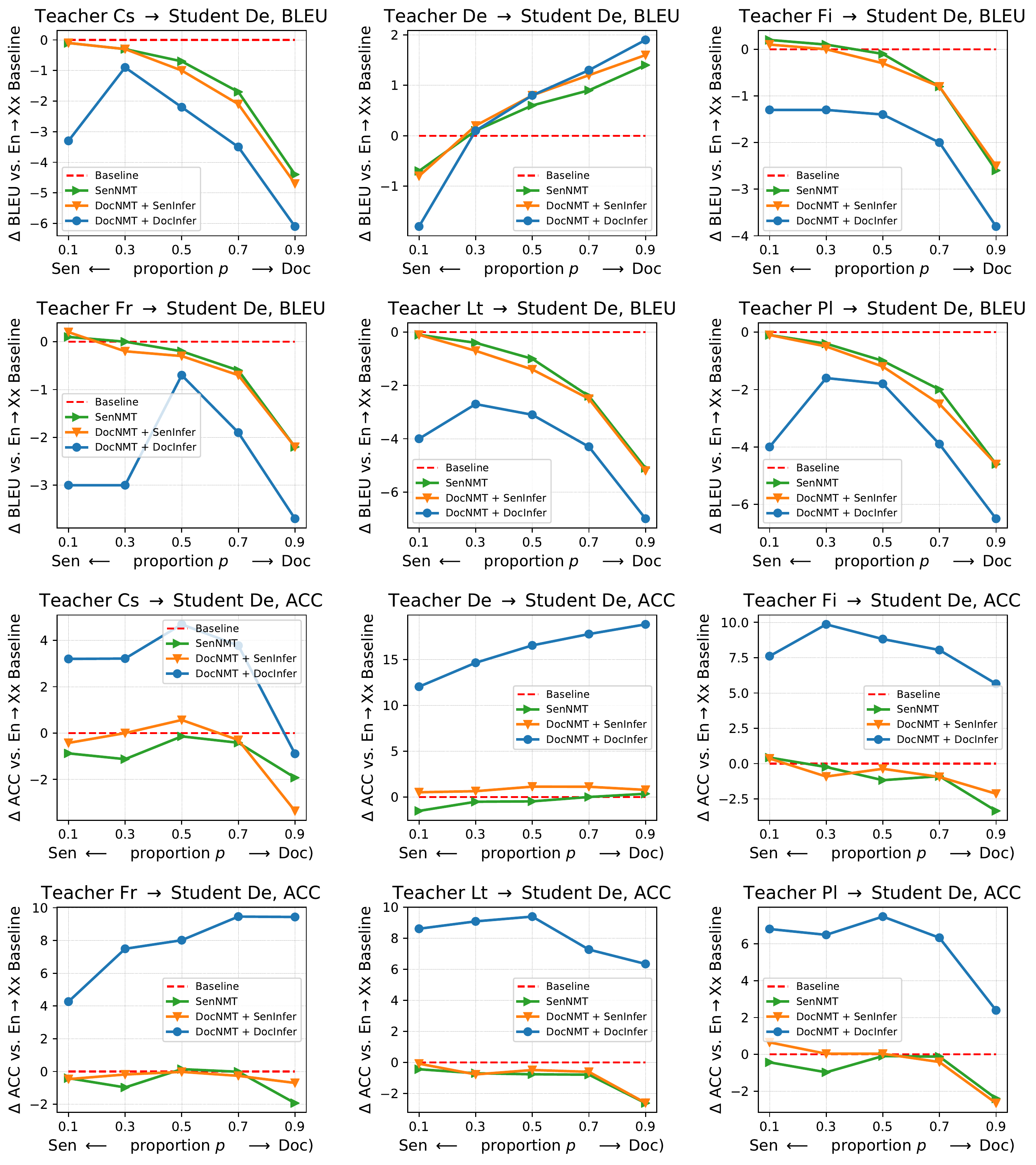}} 
    \caption{\label{fig:individual_transfer_to_de_o2m} Transfer performance from individual languages to De as a function of proportion $p$ with genuine parallel documents for \textbf{\otm} translation on \europarl.}
\end{figure*}

\begin{figure*}[ht]
    \centering
    \subfigure{\includegraphics[scale=0.43]{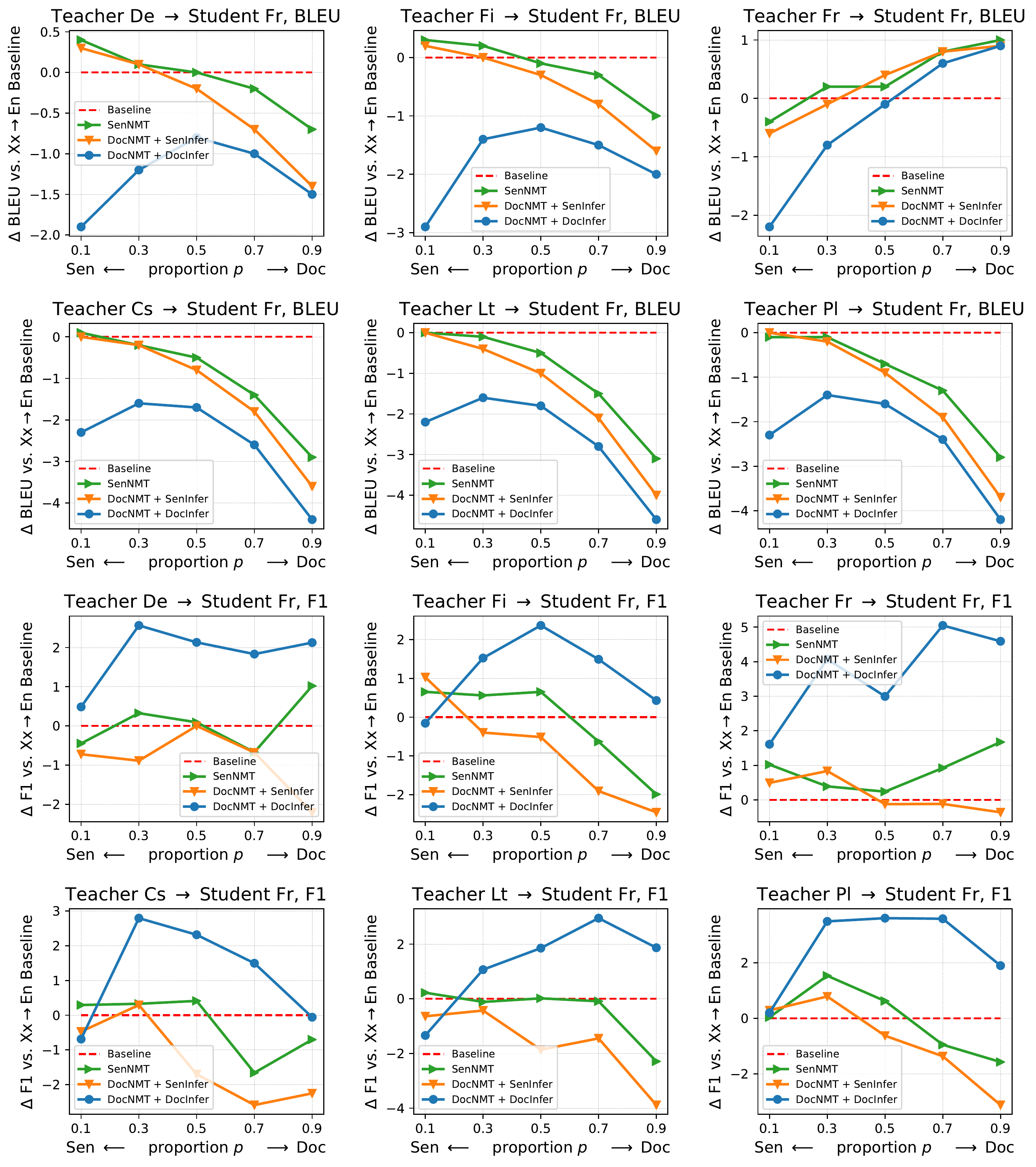}} 
    \caption{\label{fig:individual_transfer_to_fr_m2o} Transfer performance from individual languages to Fr as a function of proportion $p$ with genuine parallel documents for \textbf{\mto} translation on \europarl.}
\end{figure*}

\begin{figure*}[ht]
    \centering
    \subfigure{\includegraphics[scale=0.43]{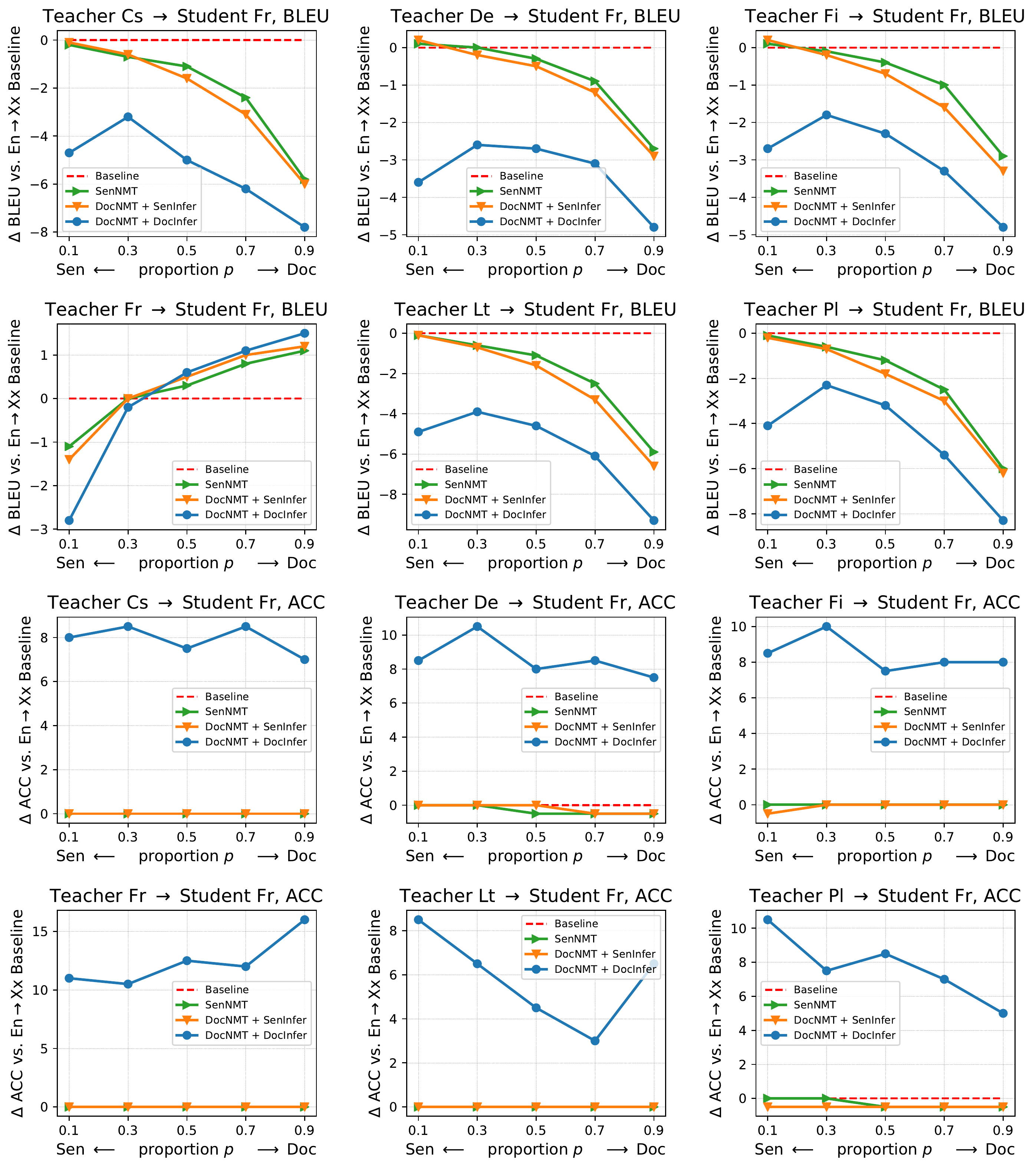}} 
    \caption{\label{fig:individual_transfer_to_fr_o2m} Transfer performance from individual languages to Fr as a function of proportion $p$ with genuine parallel documents for \textbf{\otm} translation on \europarl.}
\end{figure*}

\end{document}